\documentclass{article}

    \usepackage[preprint,nonatbib]{neurips_2020}

\usepackage[utf8]{inputenc} 
\usepackage[T1]{fontenc}    
\usepackage{hyperref}       
\usepackage{url}            
\usepackage{booktabs}       
\usepackage{amsfonts}       
\usepackage{nicefrac}       
\usepackage{microtype}      
\usepackage{color}
\usepackage{amsmath}
\usepackage{bbm}
\usepackage{booktabs}
\usepackage[ruled,vlined]{algorithm2e}
\usepackage{setspace}
\usepackage{bbold}
\usepackage{graphicx}
\usepackage{float}
\usepackage{enumitem}

\title{Learning 3D-3D Correspondences for One-shot Partial-to-partial Registration} 

\author{%
    Zheng Dang,\\
    IAIR, Xi'an Jiaotong University, China \\
	dangzheng713@stu.xjtu.edu.cn\\
	\And
	Fei Wang,\\
	IAIR, Xi'an Jiaotong University, China \\
	wfx@mail.xjtu.edu.cn\\
	\And
	Mathieu Salzmann,\\
	EPFL-CVLab \& ClearSpace, Switzerland,\\
	mathieu.salzmann@epfl.ch
}

\begin{document}

\maketitle
\newcommand{\bX}{\mathcal{X}}
\newcommand{\btX}{\tilde{\mathcal{X}}}
\newcommand{\bx}{\mathbf{x}}
\newcommand{\bbx}{\bar{\mathbf{x}}}
\newcommand{\bY}{\mathcal{Y}}
\newcommand{\by}{\mathbf{y}}
\newcommand{\bby}{\bar{\mathbf{y}}}
\newcommand{\bfx}{\mathbf{f}^x}
\newcommand{\bfy}{\mathbf{f}^y}
\newcommand{\bthx}{\theta^x}
\newcommand{\bthy}{\theta^y}
\newcommand{\bK}{\mathbf{K}}
\newcommand{\bT}{\mathcal{T}}
\newcommand{\bt}{\mathbf{t}}
\newcommand{\bS}{\mathcal{S}}
\newcommand{\bbS}{\mathcal{\bar{S}}}
\newcommand{\bP}{\mathcal{P}}
\newcommand{\bbP}{\mathcal{\bar{P}}}
\newcommand{\bM}{\mathcal{M}}
\newcommand{\bbM}{\mathcal{\bar{M}}}
\newcommand{\bA}{\mathcal{A}}
\newcommand{\bH}{\mathbf{H}}
\newcommand{\bR}{\mathbf{R}}
\newcommand{\bW}{\mathbf{W}}
\newcommand{\bU}{\mathbf{U}}
\newcommand{\bI}{\mathbf{I}}
\newcommand{\bV}{\mathbf{V}}
\newcommand{\ba}{\mathbf{a}}
\newcommand{\bb}{\mathbf{b}}
\newcommand{\bg}{\mathbf{g}}

\newcommand{\MS}[1]{{\color{green}{\bf MS: #1}}}
\newcommand{\ms}[1]{{\color{green} #1}}
\newcommand{\ZD}[1]{{\color{blue}{\bf ZD: #1}}}
\newcommand{\zd}[1]{{\color{blue}{#1}}}
\begin{abstract}
While 3D-3D registration is traditionally tacked by optimization-based methods, recent work has shown that learning-based techniques could achieve faster and more robust results. In this context, however, only PRNet can handle the partial-to-partial registration scenario. Unfortunately, this is achieved at the cost of relying on an iterative procedure, with a complex network architecture.
Here, we show that learning-based partial-to-partial registration can be achieved in a one-shot manner, jointly reducing network complexity and increasing registration accuracy. To this end, we propose an Optimal Transport layer able to account for occluded points thanks to the use of outlier bins. The resulting OPRNet framework outperforms the state of the art on standard benchmarks, demonstrating better robustness and generalization ability than existing techniques.

\end{abstract}
\section{Introduction}

Registration aims to determine the rigid transformation, i.e., 3D rotation and 3D translation, between two 3D point sets. The traditional approach to this problem is the Iterative Closest Point (ICP) algorithm~\cite{Besl92}. In its vanilla version, this iterative algorithm easily gets trapped in poor local optima. While globally-optimal solutions~\cite{Yang15,Zhou16} have been proposed to remedy this, they come at a high computational cost 
and lack robustness to noise, thus greatly reducing their practical applicability.

Recently, learning-based methods~\cite{Aoki19,Wang19e} were shown to outperform the previous traditional, optimization-based techniques in terms of both speed and robustness to noise. In particular, DCP~\cite{Wang19e} provides a one-shot strategy that prevents the need for an expensive iterative process. Unfortunately, these methods cannot handle the partial-to-partial registration scenario, thus only working under the unrealistic assumption that both point sets are fully observed, even at test time. This was addressed by PRNet~\cite{Wang19f}, but by falling back to an iterative keypoint-matching procedure, relying on a reinforcement learning-inspired actor-critic strategy. This drastically increased the complexity of the resulting network, requiring four 32GB GPUs for training, instead of just one 
for DCP.

In this paper, we introduce a more direct, one-shot learning-based solution to the partial-to-partial registration problem. To this end, we revisit the DCP framework so as to remove its assumption that each 3D point in the target (observed) point set can always be matched to a source (model) point. Specifically, we replace the softmax operator used in DCP, which not only assumes that each target point corresponds to a source point but also incorrectly allows for one-to-many correspondences, with an Optimal Transport (OT) layer. This layer leverages a measure of similarity between the target and source points to compute a transportation cost, which we then use to obtain a matching with Sinkhorn's algorithm. To model \emph{partial} correspondences, we introduce outlier bins, allowing us to account for occluded points. We further show that imposing direct supervision on the output of the OT layer, instead of computing a loss function on the rigid transformation obtained via SVD, as done in DCP, improves the robustness of the resulting model.

Our contributions can be summarized as follows:
\begin{itemize}[leftmargin=*]
\vspace{-0.15cm}
\item We introduce the first one-shot learning-based approach for partial-to-partial registration, which we dub OPRNet, for One-shot Partial-to-partial Registration Network.
\vspace{-0.05cm}
\item We develop an Optimal Transport layer able to handle partial matches thanks to outlier bins.
\vspace{-0.05cm}
\item We show that better robustness can be achieved by direct supervision of the OT layer output than by supervising the rigid transformation obtained after SVD.
\end{itemize}
\vspace{-0.15cm}
Our experiments show that, despite being a one-shot algorithm, our approach outperforms the state-of-the-art iterative PRNet on the standard ModelNet40 benchmark, including in challenging scenarios with unseen point-clouds and unseen categories at test time. Furthermore, our approach can handle high missing-point rates, such as those observed when acquiring point sets with depth sensors, and generalizes well to real data without any re-training or fine-tuning. Finally, our network retains the relatively compact size of DCP, requiring training resources around 6 times smaller than PRNet. We will release our code upon acceptance to facilitate reproducibility and future research.

\vspace{-0.0cm}
\section{Related Work}
\vspace{-0.2cm}
\textbf{Traditional point cloud registration.} ICP is the best-known algorithm for solving the point cloud registration problem. It comprises two steps: One whose goal is to find the closest target point for each source point to generate 3D-3D correspondences, and the other that computes the rigid transformation from these correspondences by solving a least-square problem. These two steps are repeated until a termination condition is satisfied. Several variants, such as Generalized-ICP~\cite{Segal09} and Sparse ICP~\cite{Bouaziz13}, have been proposed to improve  robustness to noise and mismatches, and we refer the reader to~\cite{Pomerleau15,Rusinkiewicz01} for a complete review of ICP-based strategies. The main drawback of these methods is their requirement for a reasonable initialization to converge to a good solution. Only relatively recently has this weakness been addressed by the globally-optimal registration method Go-ICP~\cite{Yang15}. In essence, this approach follows a branch-and-bound strategy to search the entire 3D motion space $SE(3)$. Motivated by this, other approaches to finding a global solution have been proposed, via, e.g., Riemannian optimization~\cite{Rosen19}, convex relaxation~\cite{Maron16}, and mixed-integer programming~\cite{Jzatt20}. While globally optimal, these methods all come at a much higher computational cost than vanilla ICP. This was, to some degree, addressed by the Fast Global Registration (FGR) algorithm~\cite{Zhou16}, which leverages a 
local refinement strategy to speed up computation. While effective, FGR still suffers from the presence of noise and outliers in the point sets, particularly because, as vanilla ICP, it simply relies on 3D point-to-point distance to establish correspondences. In principle, this can be addressed by designing point descriptors that can be more robustly matched. Over the years, several works have tackled this task, in both a non learning-based~\cite{Johnson99,Rusu08,Rusu09} and learning-based~\cite{Zeng17b,Khoury17} fashion. Nowadays, however, these approaches are outperformed by end-to-end learning frameworks, which directly take the point sets as input, as discussed below.

\textbf{End-to-end learning with point sets.} A key requirement to enable end-to-end learning-based registration was the design of deep networks acting on unstructured sets. Deep sets~\cite{Zaheer17} and PointNet~\cite{Qi17} constitute the pioneering works in this direction. They  use shared multilayer perceptrons to extract high-dimensional features from the input point coordinates, and exploit a symmetric function to aggregate these features. This idea was then extended in PointNet++~\cite{Qi17a} via a modified  sampling strategy to robustify the network to point clouds of varying density, in DGCNN~\cite{Wang18b} by building a graph over the point cloud, in PointCNN~\cite{Li18c} by learning a transformation of the data so as to be able to process it with standard convolutional layers, and in PCNN~\cite{Atzmon18} via an additional  extension operator before applying the convolutions. While the above-mentioned works focused on other tasks than us, such as point cloud classification or segmentation, end-to-end learning for registration has recently attracted a growing attention. In particular, PointNetLK~\cite{Aoki19} combines the PointNet backbone with the traditional, iterative Lucas-Kanade (LK) algorithm~\cite{Lucas81} so as to form an end-to-end registration network; DCP~\cite{Wang19e} exploits DGCNN backbones followed by Transformers~\cite{Vaswani17} to establish 3D-3D correspondences, which are then passed through an SVD layer to obtain the final rigid transformation. While effective, PointNetLK and DCP cannot tackle the partial-to-partial registration scenario. That is, they assume that both point sets are fully observed, during both training and test time. This was addressed by PRNet~\cite{Wang19f} via a deep network designed to extract keypoints from each input set and match these keypoints. This network is then applied in an iterative manner, so as to increasingly refine the resulting transformation. Unfortunately, this iterative procedure, and the corresponding actor-critic matching strategy, significantly increase the network resource requirements compared to the one-shot DCP. Here, we show that accurate, one-shot partial-to-partial registration can be achieved with a network requiring resources similar to DCP.

\section{Methodology}
\subsection{Problem Formulation}
Let $\bX \in \mathbb{R}^{M \times 3}$ and $\bY \in \mathbb{R}^{N \times 3}$ be two sets of 3D points sampled from the same object surface. We typically refer to $\bX$ as the source point set and to $\bY$ as the target point set. Registration then aims to find a rigid transformation $\bT$ that aligns $\bX$ to $\bY$. In this work, we tackle the partial-to-partial scenario, in which both sets of samples can vary significantly, but nonetheless have \emph{some} overlap.
 
\begin{figure}[t]
    \centering
    \setlength{\belowcaptionskip}{-10pt}
    \includegraphics[width=11cm]{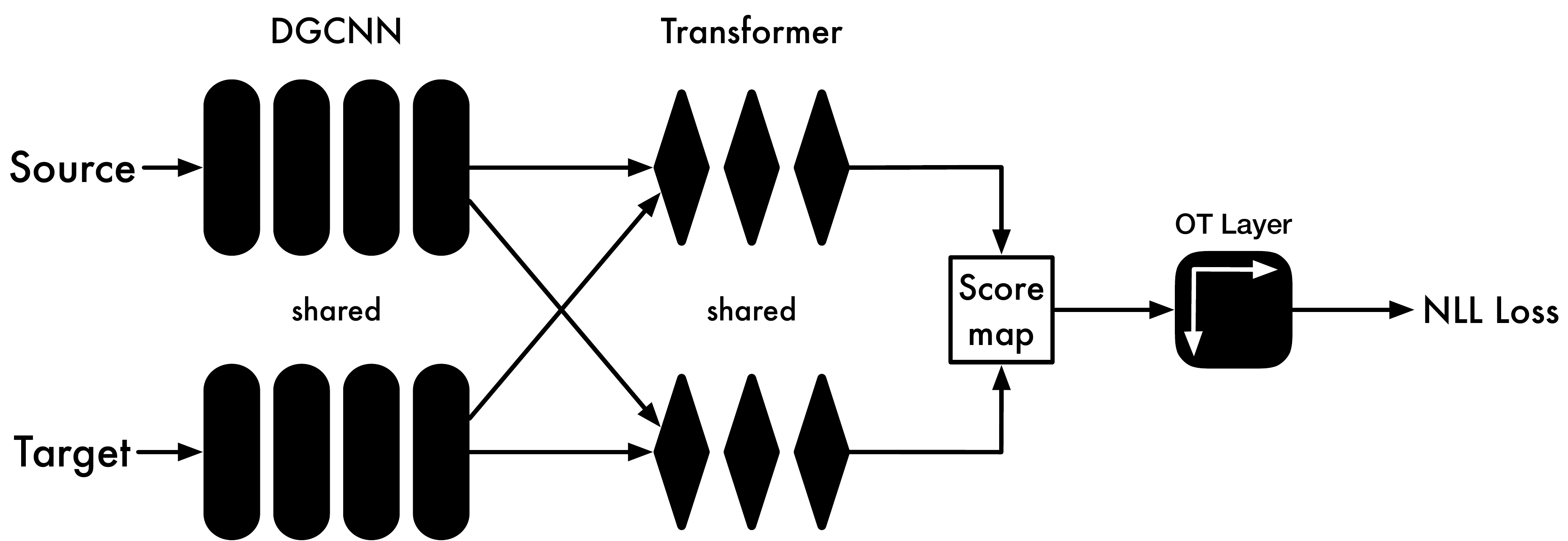}
    \caption{\label{image1} Network architecture of OPRNet.}
\end{figure}

As discussed in~\cite{Besl92}, given correspondences between the two point sets, the rigid transformation $\bT$ can be obtained in closed form by solving a Procrustes problem. Here, we therefore develop a one-shot deep network that produces such correspondences. To this end, as illustrated by Fig.~\ref{image1}, we make use of a feature embedding backbone that extracts a representation for each point in each set, and compute an $M\times N$ score map $\bS$ from these features. We then formulate the matching problem as an optimal transport~\cite{Villani08} one, for which we obtain an approximate solution via a log-domain Sinkhorn algorithm~\cite{Cuturi13} incorporated in the network. Below, we discuss these steps in more detail.

\subsection{Feature Embedding Backbone}
To process the individual point sets, we make use of the same DGCNN~\cite{Wang18b} + Transformer~\cite{Vaswani17} combination as in DCP(v2)~\cite{Wang19e}. 
Specifically, 
DGCNN takes a point set as input, from which it constructs a k-NN graph. It then extracts point-wise features via standard convolutions on this graph, encoding diverse levels of context by max-pooling the local features and concatenating the resulting representations to the point-wise ones.
Let $\bthx$, resp. $\bthy$, be the final feature matrix, i.e., one $P$-dimensional feature vector per 3D point, for $\bX$, resp. $\bY$. The transformer then learns a function $\phi : \mathbb{R}^{M\times P} \times \mathbb{R}^{N\times P} \rightarrow \mathbb{R}^{M\times P}$, that combines the information of the two point sets. 
Ultimately, this backbone produces descriptor matrices $\bfx$,
resp. $\bfy$,
for $\bX$, resp. $\bY$, as
\begin{equation}
    \bfx = \bthx + \phi(\bthx, \bthy), \;\;\;
    \bfy = \bthy + \phi(\bthy, \bthx)\;.
\end{equation}

Given these matrices, we then form a score map $\bS \in \mathbb{R}^{M\times N}$ by computing the similarity between each source-target pair of descriptors. That is, we compute the $(i,j)$-th element of $\bS$ as
\begin{equation}
    \bS_{i,j} = <\bfx_i, \bfy_j>, \;\;\forall (i, j) \in [1,M] \times [1,N]\;,
\end{equation}
where $<\cdot, \cdot>$ is the inner product, and $\bfx_i, \bfy_j \in \mathbb{R}^P$. 
In DCP~\cite{Wang19e}, this score map is passed through a row-wise softmax so as to obtain correspondences. These correspondences are then processed within the network via an SVD layer to solve the Procrustes problem, and the resulting rigid transformation is compared to the ground-truth one with a mean squared error (MSE) loss. 

This approach suffers from the following drawbacks. First, the softmax operator assumes that every target point matches a source point, which does not hold in partial-to-partial registration. Second, it allows multiple target points to correspond to the same source point. Third, in practice, the use of an SVD layer, if not carefully designed, can make the training process unstable~\cite{Dang18,Dang20,Wang19g}. 
To address this, below, we introduce an OT layer that explicitly reflects the underlying correspondence problem, and enable it to handle partially-observed point sets via outlier bins. Furthermore, we improve training stability by imposing direct supervision on the output of the OT layer.

\subsection{Optimal Transport Layer}
Given the score matrix $\bS \in \mathbb{R}^{M \times N}$, we seek to obtain a partial assignment of the target points to the source points, which would effectively provide us with correspondences. In this section, we first introduce the outlier bins that allow us to handle partial point sets, and then discuss how we compute the partial assignment.

\textbf{Outlier bins.} 
Because we tackle partial-to-partial registration, some points in one set will have no corresponding point in the other. To model this, inspired by~\cite{Detone18},
which tackles the 2D-2D matching scenario, we adopt a dustbin design. Specifically, we extend the score matrix $\bS$ by one row and one column so as to form an augmented score matrix $\bbS$. The values at the newly-created positions in $\bbS$ are then set to
\begin{equation}
    \bbS_{i, N + 1} = \bbS_{M + 1, j} = \bbS_{M + 1, N + 1} = \alpha, \;\;\forall i \in [1, M], \;\forall j \in [1, N]\;,
 \end{equation}
where $\alpha \in \mathbb{R}$ is a learnable parameter. The values at the other indices directly come from $\bS$.

\textbf{Optimal transport.} 
Given the augmented score map $\bbS$, we aim to find a partial assignment $\bbP \in \mathbb{R}^{(M+1)\times (N+1)}$, defining correspondences between the two point sets, extended with the outlier bins. For such an assignment to be valid, it should ensure that each point in one set has at most one corresponding point in the other. However, it should allow multiple points to be assigned to the outlier bins. We express this by constraining $\bbP$ to belong to the set
\begin{equation}
    \bU(\ba, \bb) = \left\{\bbP \in \mathbb{R}_{+}^{(M+1)\times (N+1)}: \bbP \mathbbm{1}_{M+1} = \ba \:\text{and}\: \bbP^{\top} \mathbbm{1}_{N+1} = \bb \right\}\;,
\end{equation}
where $\ba = [\mathbbm{1}_{M}^{\top}, N]^{\top}$, and $\bb = [\mathbbm{1}_{N}^{\top}, M]^{\top}$, with $\mathbbm{1}_{M} = [1, 1, ..., 1]^{\top}\in \mathbb{R}^{M}$. Note that, here, we further make use of Kantorovich's relaxation, which allows the elements of $\bbP$ to take any non-negative value instead of constraining them to be in $\{0,1\}$. More detail on this relaxation can be found in Section 2.3 of~\cite{Cuturi13}.
The resulting set $\bU(\ba, \bb)$ is bounded and defined by $N + M + 2$ equality constraints, forming a convex polytope.


Then, from the optimal transport theory~\cite{Peyre19,Cuturi13}, the probability matrix $\bbP$ can be obtained by solving
\begin{equation}
\min_{\bbP \in \bU (\ba, \bb)} \langle \bbS, \bbP \rangle - \lambda E(\bbP)\;,
\label{entropy_func}
\end{equation}
where $\langle\cdot,\cdot\rangle$ is the Frobenius dot product and 
$E(\cdot)$ is an entropy regularization term defined as $E(\bbP) = - \displaystyle\sum_{i, j} \bbP_{i, j} \left(\log(\bbP_{i, j}) - 1\right)$. 

\textbf{Log-domain Sinkhorn’s algorithm.} 
The standard way to solve~\eqref{entropy_func}
consists of making use of Sinkhorn's algorithm~\cite{Sinkhorn67,Cuturi13}. In concurrent work, this algorithm has been employed within deep networks for 2D-2D matching with a similar dustbin design to ours~\cite{Sarlin19}, and for 2D-3D matching assuming that every point has a correspondence~\cite{Liu20}. 
In practice, we found that using the original version of this algorithm within a deep network, as in~\cite{Liu20}, suffers from numerical overflow, preventing training to converge. To address this, we adopt the log-domain computation suggested in Section 4.4 of~\cite{Cuturi13}. 
The resulting algorithm is provided in Algorithm~\ref{alg:sinkhorn}, where the matrix operator $logsumexp(\bA) = log(exp(\bA_{1,1}) + ... + exp(\bA_{i,j}) + ... + exp(\bA_{M,N}))$.
Since all operations performed by this algorithm are differentiable, the training errors can be backpropagated to the rest of the network.

\begin{algorithm}[H]
    \setstretch{1.35}
    \SetAlgoLined
    \SetKwInOut{Input}{Input}
    \SetKwInOut{Output}{Output}
    \SetKwInOut{Init}{Init}
    \SetKwComment{Comment}{$\triangleright$\ }{}
    \Input{
        $\bbS$, $\ba=[\mathbbm{1}_{M}^{\top}, N]^{\top}$, $\bb=[\mathbbm{1}_{N}^{\top}, M]^{\top}$, $\mathbf{f}_{0}=\mathbb{0}_{M+1}$, $\bg_{0}=\mathbb{0}_{N+1}$, $\lambda$ and $k$ (nb. of iters)
        }
    \Output{
        Assignment matrix $\bbP$
        }
    \Init{
        $\bbS = -\bbS / \lambda$
        }
    \While{$l \leq k$}{
        $\mathbf{f}^{(l + 1)} = \lambda log(\ba) - \mathit{logsumexp}(\bbS + \mathbbm{1}_{N+1}\bg^{(l)\top})$ \\
        $\bg^{(l + 1)} = \lambda log(\bb) - \mathit{logsumexp}(\bbS + \mathbf{f}^{(l+1)}\mathbbm{1}_{M+1}^{\top})$ 
    }
    $\bbP_{i,j}=\mathit{exp}(\bbS_{i,j} + \mathbf{f}_i + \mathbf{g}_j)$ \Comment*[r]{transfer back to the original domain.}
    \caption{Log-domain Sinkhorn's algorithm.}
    \label{alg:sinkhorn}
    
\end{algorithm}

\subsection{Loss Function}

To train our network, we propose to rely on direct supervision of the output of the OT layer. To this end, let $\bbM \in \{0, 1\}^{(M+1) \times (N+1)}$ be the matrix of ground-truth correspondences, with a 1 indicating a correspondence between a pair of points. Such correspondences can be estimated using the ground-truth transformation matrix, as discussed in more detail in Section~\ref{build_gt}. We then express our loss function as the negative log-likelihood of the assignment, given by
\begin{equation}
    \mathcal{L}(\bbP, \bbM) = \frac{- \sum\limits_{i = 1}^{M + 1}\sum\limits_{j = 1}^{N + 1} (\log \bbP_{i,j})\bbM_{i, j}}{\sum\limits_{i = 1}^{M + 1}\sum\limits_{j = 1}^{N + 1} \bbM_{i, j}}\;,
\end{equation}
where the denominator normalizes the loss value so that different training samples containing different number of correspondences have the same influence in the overall empirical risk.

\subsection{Obtaining the Rigid Transformation at Test Time}
At test time, we compute correspondences from the output $\bbP$ of the OT layer. To this end, we search for the largest value in each row of $\bbP$. If the resulting index is a valid point, we treat the pair as a correspondence; if the largest value occurs in the outlier bin, we then discard the current sample. 
With the resulting set of matches, we obtain the rigid transformation by solving the Procrustes problem. We refer the reader to the supplementary material for the details of this step.

\section{Experiments}
\begin{table}[t]
    \centering
    \setlength{\abovecaptionskip}{0.3cm}
    \setlength{\belowcaptionskip}{-0pt}
    \setlength{\tabcolsep}{3.25mm}{
    \begin{tabular}{@{}lllllll@{}}
    \toprule
    \textbf{Model}      & \textbf{MSE(R)} & \textbf{RMSE(R)} & \textbf{MAE(R)} & \textbf{MSE(t)} & \textbf{RMSE(t)} & \textbf{MAE(t)} \\ \midrule
    \textbf{ICP}        & 1134.552        & 33.683           & 25.045          & 0.0856          & 0.293            & 0.250           \\
    \textbf{FGR}        & 126.288         & 11.238           & 2.832           & 0.0009          & 0.030            & 0.008           \\
    \textbf{Go-ICP}     & 195.985         & 13.999           & 3.165           & 0.0011          & 0.033            & 0.012           \\
    \textbf{PointNetLK} & 280.044         & 16.735           & 7.550           & 0.0020          & 0.045            & 0.025           \\
    \textbf{DCP(v2)}    & 45.005          & 6.709            & 4.448           & 0.0007          & 0.027            & 0.020           \\
    \textbf{PRNet}      & 10.235          & 3.199            & 1.454           & 0.0003          & 0.016            & 0.010           \\ \midrule
    \textbf{Ours}       & \textbf{0.107}  & \textbf{0.328}   & \textbf{0.0521} & \textbf{3.384e-06} & \textbf{0.00183} & \textbf{0.000281}        \\ \bottomrule
    \end{tabular}}
    \caption{\label{table1} Partial-to-partial registration with unseen objects.}
\end{table}

\begin{table}[t]
    \setlength{\abovecaptionskip}{0.2cm}
    \setlength{\belowcaptionskip}{-10pt}
    \centering
    \setlength{\tabcolsep}{3.25mm}{
    \begin{tabular}{@{}lllllll@{}}
    \toprule
    \textbf{Model}      & \textbf{MSE(R)} & \textbf{RMSE(R)} & \textbf{MAE(R)} & \textbf{MSE(t)} & \textbf{RMSE(t)} & \textbf{MAE(t)} \\ \midrule
    \textbf{ICP}        & 1217.618        & 34.894           & 25.455          & 0.086           & 0.293            & 0.251           \\
    \textbf{FGR}        & 98.635          & 9.932            & 1.952           & 0.0014          & 0.038            & 0.007           \\
    \textbf{Go-ICP}     & 157.072         & 12.533           & 2.940           & 0.0009          & 0.031            & 0.010           \\
    \textbf{PointNetLK} & 526.401         & 22.943           & 9.655           & 0.0037          & 0.061            & 0.033           \\
    \textbf{DCP(v2)}    & 95.431          & 9.769            & 6.954           & 0.0010          & 0.034            & 0.025           \\
    \textbf{PRNet}      & 24.857          & 4.986            & 2.329           & 0.0004          & 0.021            & 0.015           \\
    \textbf{PRNet*}     & 15.624          & 3.953            & 1.712           & 0.0003          & 0.017            & 0.011           \\ \midrule
    \textbf{Ours}       & \textbf{0.127}  & \textbf{0.357}   & \textbf{0.0696} & \textbf{3.953e-06} & \textbf{0.00198} & \textbf{0.000406}         \\ \bottomrule
    \end{tabular}}
    \caption{\label{table2} Partial-to-partial registration with unseen categories.}
\end{table}
We first compare our approach to the state of the art in the same settings used in~\cite{Wang19e,Wang19f}, and then evidence its effectiveness in the more challenging scenario where self-occluded points in one set are discarded, as with, e.g., a LiDAR or depth sensor, and on real scans. Finally we provide an ablation study of the different components of our approach.
\subsection{Partial-to-Partial Registration on ModelNet40}
    \label{modelnet40}

    \textbf{ModelNet40.} This dataset~\cite{Wu15} contains 12,311 CAD models spanning 40 object categories. As in~\cite{Wang19f}, we split it into 9,843 training and 2,468 testing samples. The point clouds in this dataset are normalized in the range [-1, 1] on each axis. During training, we sample 1024 points from the original point clouds to define the source point set $\bX$. We then sample a random rigid transformation along each axis, with rotation in $[0^{\circ}, 45^{\circ}]$ and translation in $[-0.5, 0.5]$, and apply it to the point set $\bX$ to obtain the target point set $\bY$. To simulate partial scans of $\bX$ and $\bY$, as in~\cite{Wang19f}, we randomly select one point in $\bX$ and one point in $\bY$ independently, and only keep their 768 nearest neighbors.
    
    \textbf{Evaluation metrics.} We report the mean squared error (MSE), root mean squared error (RMSE), and mean absolute error (MAE) between the predicted transformation and the ground truth, treating rotation and translation separately. For the rotation, we evaluate these errors in degrees. 
    
    \textbf{Implementation details.} We implemented our network using Pytorch~\cite{Paszke17} and trained it from scratch. We used the Adam optimizer~\cite{Kingma15} and set the learning rate to $10^{-3}$. We relied on mini-batches of size 20 and trained the network for 40,000 iterations. For the OT layer, we used $k=50$ iterations and set $\lambda=1$. Training was performed on two NVIDIA 2080Ti GPUs.
    
    \label{build_gt}
    To build the ground-truth assignment matrix $\bbM$, we transformed $\bX$ using the ground-truth transformation $\bT$ to get a transformed point set $\btX$. We then computed the pair-wise Euclidean distance matrix between $\btX$ and $\bY$, which we thresholded to 0.05 to obtain a correspondence matrix $\bM \in \{0, 1\}$. Then we augmented $\bM$ with an extra row and column acting as outlier bins to get $\bbM$. The points without any correspondence were treated as outliers, and the corresponding positions in $\bbM$ were set to one. While this strategy does not guarantee a  bipartite matching, in practice, we found that one-to-many correspondences occurred only very rarely. Strict bipartite matchings could be obtained via forward-backward check, but we believe that the benefit of such an operation would be outweighed by its computational cost.
    
    \textbf{Partial-to-Partial registration with unseen objects.} As a first experiment, we tackle the case where all categories are observed during training, but new objects from these categories are provided at test time. To this end, we train on all 9,843 samples and test on the remaining 2,468 ones, as in~\cite{Wang19f}. In Table~\ref{table1}, we compare our OPRNet with ICP, FGR~\cite{Zhou16}, PointNetLK~\cite{Aoki19}, DCP(v2)~\cite{Wang19e} and PRNet~\cite{Wang19f}. OPRNet outperforms all the competitors by a large margin in all the metrics.
    
    \textbf{Partial-to-Partial registration with unseen categories.} To test the generalization ability of our method, we perform an unseen-categories experiment similar to that in~\cite{Wang19f}. Specifically, we split the ModelNet40 objects into separate training and testing categories. However, instead of training our model on ShapeNetCore~\cite{Chang15} and testing it on the resulting ModelNet40 testing data, as done in~\cite{Wang19f} for PRNet, we restrict ourselves to the much smaller amount of ModelNet40 training data. For the comparison to be fair, we thus report the results of~\cite{Wang19f} trained on the same data as us as \emph{PRNet}, and that of the model trained on ShapeNetCore as \emph{PRNet*}. In any event, as shown in Table~\ref{table2}, our OPRNet outperforms all the baselines, whether trained on the same data as us or on a larger dataset. 
    
    \begin{table}[t]
    \setlength{\abovecaptionskip}{0.3cm}
    \setlength{\belowcaptionskip}{-7pt}
    \centering
    \setlength{\tabcolsep}{3.25mm}{
    \begin{tabular}{@{}lllllll@{}}
    \toprule
    \textbf{Model}      & \textbf{MSE(R)} & \textbf{RMSE(R)} & \textbf{MAE(R)} & \textbf{MSE(t)} & \textbf{RMSE(t)} & \textbf{MAE(t)} \\ \midrule
    \textbf{ICP}        & 1229.670        & 35.067           & 25.564          & 0.0860          & 0.294            & 0.250           \\
    \textbf{FGR}        & 764.671         & 27.653           & 13.794          & 0.0048          & 0.070            & 0.039           \\
    \textbf{Go-ICP}     & 150.320         & 12.261           & 2.845           & 0.0008          & 0.028            & 0.029           \\
    \textbf{PointNetLK} & 397.575         & 19.939           & 9.076           & 0.0032          & 0.0572           & 0.032           \\
    \textbf{DCP(v2)}    & 47.378          & 6.883            & 4.534           & 0.0008          & 0.028            & 0.021           \\
    \textbf{PRNet}      & 18.691          & 4.323            & 2.051           & 0.0003          & 0.017            & 0.012           \\ \midrule
    \textbf{Ours}       & \textbf{4.232}  & \textbf{2.057}   & \textbf{0.677}  & \textbf{7.145e-05} & \textbf{0.00845} & \textbf{0.00270}         \\ \bottomrule
    \end{tabular}}
    \caption{\label{table3} Partial-to-partial registration with unseen point clouds containing Gaussian noise.}
\end{table}
    \begin{table}[t]
    \centering
    \setlength{\abovecaptionskip}{0.4cm}
    \setlength{\belowcaptionskip}{-5pt}
    \setlength{\tabcolsep}{2.9mm}{
    \begin{tabular}{@{}lllllll@{}}
    \toprule
    \textbf{Model}      & \textbf{MSE(R)} & \textbf{RMSE(R)} & \textbf{MAE(R)} & \textbf{MSE(t)} & \textbf{RMSE(t)} & \textbf{MAE(t)} \\ \midrule
    \textbf{ICP*}       & 1152.602        & 33.950           & 19.651          & 0.0208          & 0.144            & 0.116           \\
    \textbf{FGR*}       & 25.890          & 5.088            & 1.275           & 0.00183         & 0.0428           & 0.0109          \\
    \textbf{DCP(softmax)}    & 0.796          & 0.892            & 0.151           & 5.307e-05          & 0.00728            & 0.00134           \\ \midrule
    \textbf{Ours}       & \textbf{0.116} & \textbf{0.341}    & \textbf{0.0902} & \textbf{7.369e-06} & \textbf{0.00271} & \textbf{0.000790}        \\ \bottomrule
    \end{tabular}}
    \caption{\label{table4} Registration with self-occluded point sets. Note that ICP and FGR failed in several cases, leading to unreasonably high errors. The * therefore indicates that we report the average over the cases where MSE < 3000. 
    }
\end{table}
    \textbf{Partial-to-Partial registration with unseen objects with gaussian noise.} To study the sensitivity of our method to noise, we perform an experiment in the same setting as the first one, but adding Gaussian noise to the data. As in~\cite{Wang19e,Wang19f}, the noise was sampled from $\mathcal{N}(0, 0.01)$ and clipped to $[-0.05, 0.05]$. 
    As shown in Table~\ref{table3}, our method still outperform all the competitors.

\subsection{Registration with Self-occluded Point Sets}
    We now turn to the case where we account for self-occlusion in one of the point sets. This mimics the scenario where the point set is captured by a sensor such as LiDAR or a depth camera. This is particularly challenging, since, in essence, only roughly half of the points are observed. 

    \textbf{Self-occluded dataset.} For this experiment, we use the auto-aligned ModelNet40 dataset~\cite{Wu15,Sedaghat16}.  This dataset contains the 40 object categories' mesh models, which we use to render depth maps given camera viewpoints and intrinsic parameters. Specifically, we treat the full point clouds corresponding to the meshes as source sets and the resulting depth maps as target sets. We use the look-at method\footnote{https://www.scratchapixel.com/lessons/mathematics-physics-for-computer-graphics/lookat-function} to place the camera, and set the distance between the camera and the center of the object to be 3. We then randomly sample the elevation and azimuth in $[22.5^{\circ}, 67.5^{\circ}]$.

    \textbf{Implementation details.} In the training process, we render the depth maps online. 
    We set the number of source points to be 1024, and that of target points to be 512. 
    We use the same hyper-parameter values as in the previous experiments, except for the number of training iterations, which we restrict to 20,000 because online depth rendering increases training time, which we  keep below 11 hours.
    

    \begin{figure}[t]
    \setlength{\belowcaptionskip}{-10pt}
    \includegraphics[width=2.65cm]{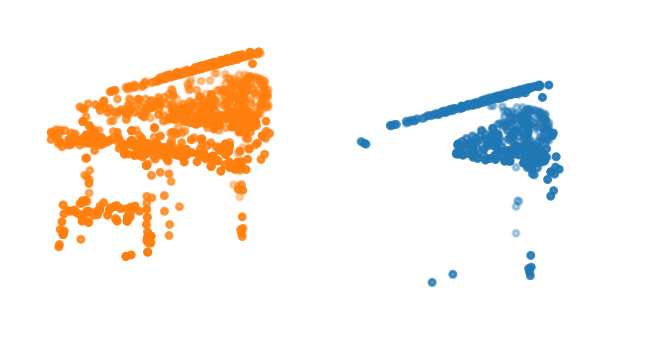}
    \includegraphics[width=2.65cm]{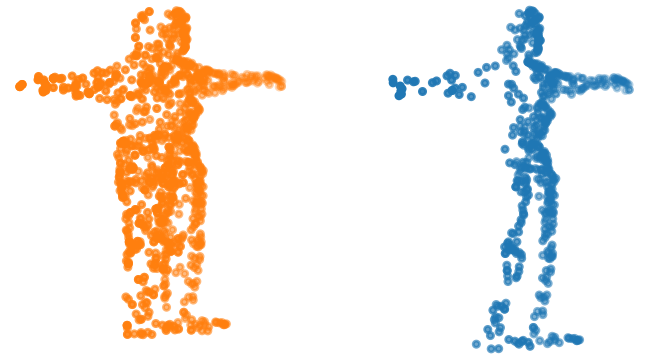}
    \includegraphics[width=2.65cm]{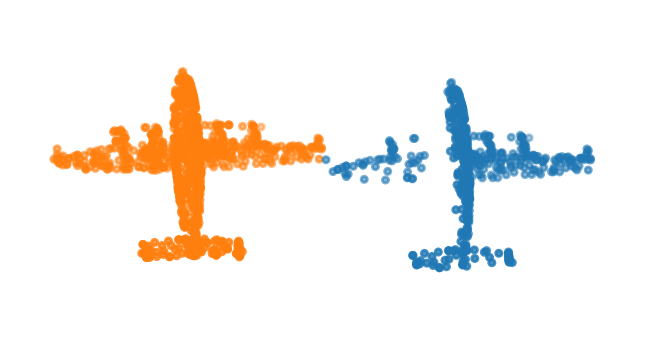}
    \includegraphics[width=2.65cm]{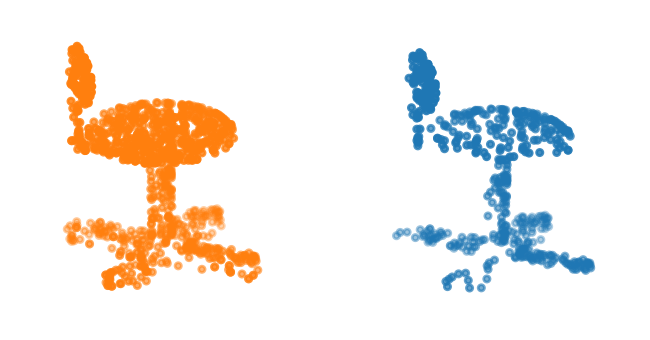}
    \includegraphics[width=2.65cm]{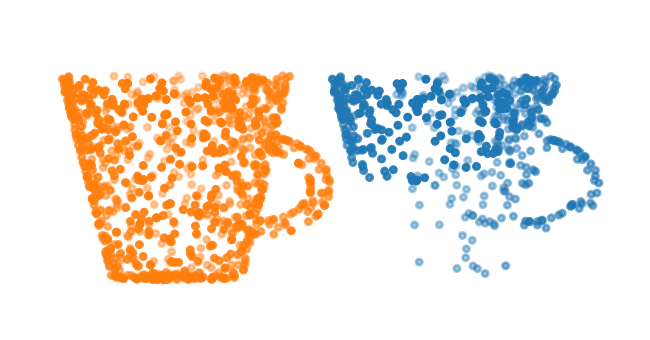}
    \includegraphics[width=2.7cm]{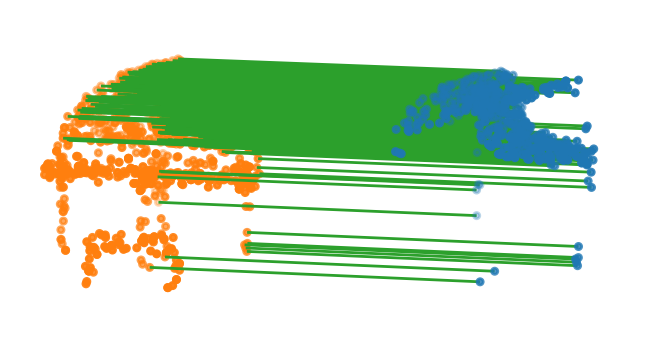}
    \includegraphics[width=2.7cm]{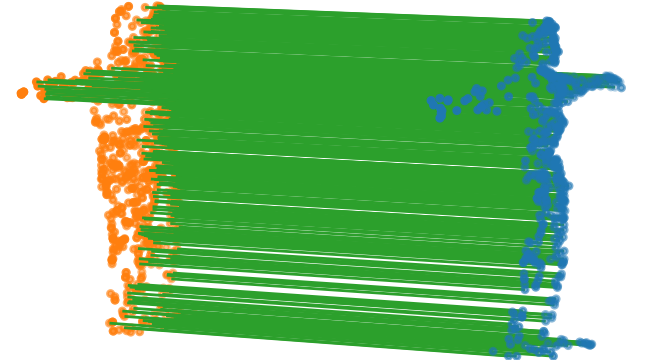}
    \includegraphics[width=2.7cm]{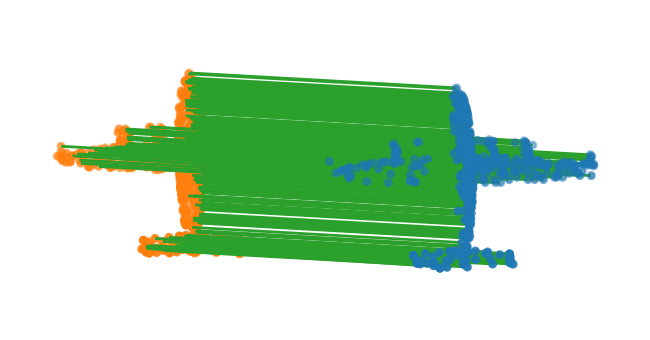}
    \includegraphics[width=2.7cm]{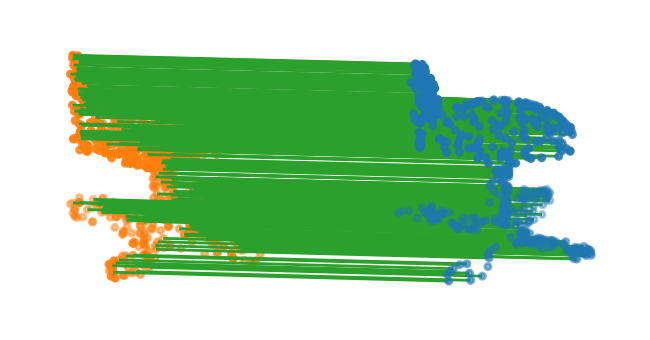}
    \includegraphics[width=2.7cm]{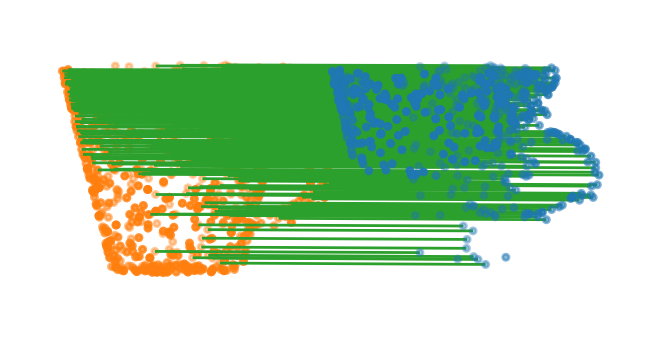}
    \caption{\label{5_obj_match} 
    Alignments and correspondences obtained with our approach in the self-occlusion scenario. We plot the point sets (source in orange, target in blue) after alignment, but shifted horizontally for clarity. The green lines in the bottom row depict the correspondences found with our approach.
    }
\end{figure}
    \textbf{Registration results.} 
    We compare our approach with ICP and FGR. However, we do not compare it against PRNet, because we do not have access to the four 32GB GPUs required to train it. Furthermore, because the rendering process was implemented using Pytorch3D, which relies on Pytorch1.5, the runtime of the SVD layer increased significantly, to the point of requiring 416 hours to train DCP. To circumvent this, we therefore replaced the SVD layer + MSE loss of DCP with a softmax layer and the same NLL loss as us, and refer to this baseline as DCP(softmax). Note that, as will be shown in Section~\ref{sec:analysis}, replacing SVD + MSE with softmax + NLL improves accuracy, and we therefore believe that our comparison is fair. The results of this comparison are provided in Table~\ref{table4}, where we again consistently outperform the competitors.
    In Fig.~\ref{5_obj_match}, we visualize qualitative results of our approach. Note that the correspondences and alignments we obtain are highly accurate.

\subsection{Partial-to-Partial Registration of Real Scans}
    \begin{figure}[t]
    \setlength{\belowcaptionskip}{-5pt}
    \includegraphics[width=2.65cm]{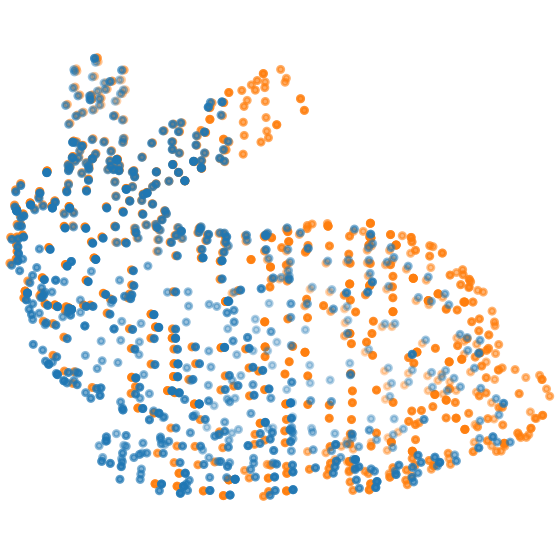}
    \includegraphics[width=2.65cm]{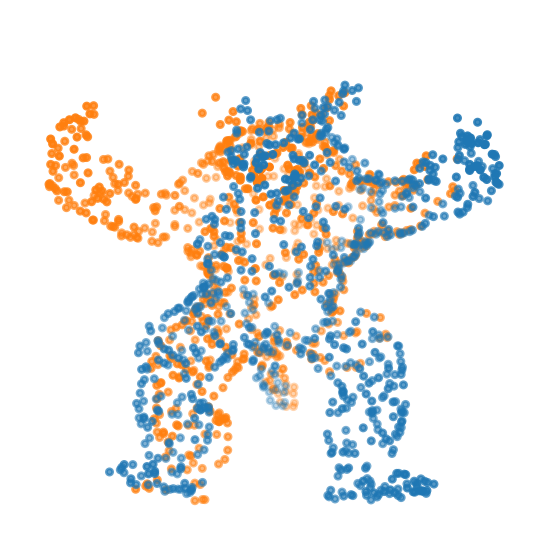}
    \includegraphics[width=2.65cm]{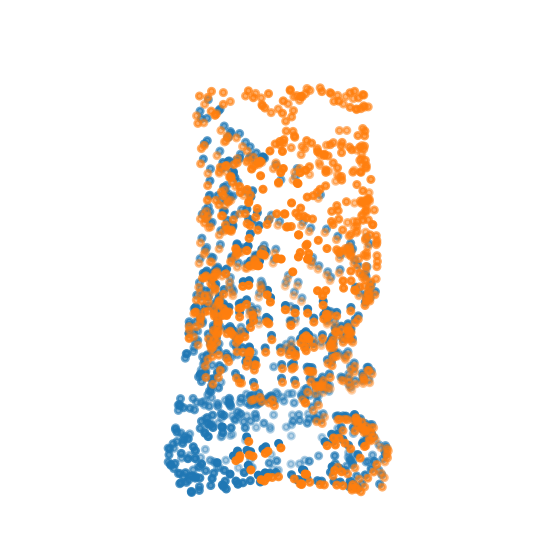}
    \includegraphics[width=2.65cm]{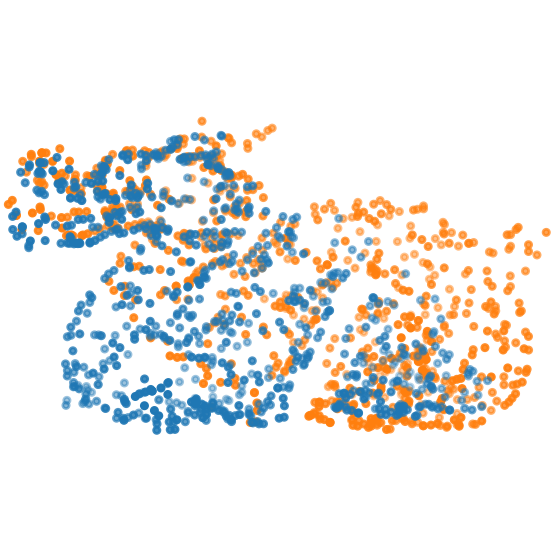}
    \includegraphics[width=2.65cm]{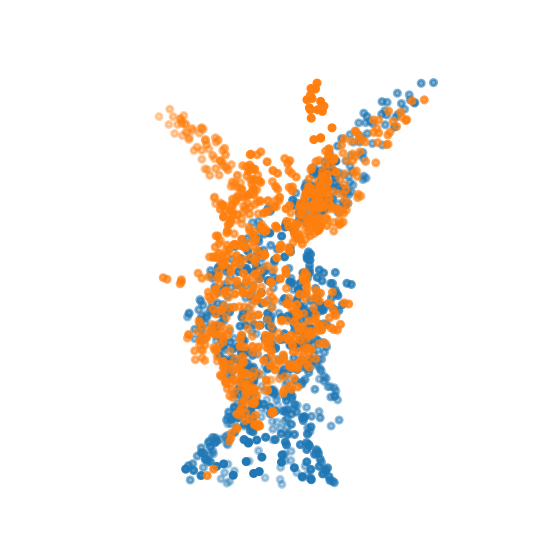}
    \caption{\label{stanford} Registering real scans. We successfully align the source (orange) and target (blue) sets.} 
\end{figure}
    We now demonstrate the use of our approach on the Stanford Bunny dataset~\cite{Turk94}, which contains real scans of objects and was also used in~\cite{Wang19f}. We process the data in the same way as in Section \ref{modelnet40}. Unlike in~\cite{Wang19f}, where PRNet had to be fine-tuned to handle such real scans, we directly apply our OPRNet trained for the ModelNet40 unseen object experiments. As shown by Fig.~\ref{stanford}, our model yields accurate alignments, thus evidencing the good generalization ability of our approach.
    
\subsection{Method Analysis}
\label{sec:analysis}
\begin{table}[t]
    \centering
    \setlength{\abovecaptionskip}{0.3cm}
    \setlength{\belowcaptionskip}{-5pt}
    \setlength{\tabcolsep}{8.8mm}{
    \begin{tabular}{@{}llllll@{}}
    \toprule
    \textbf{Points}      & \textbf{ICP} & \textbf{FGR} & \textbf{DCP} & \textbf{PRNet} & \textbf{Ours} \\ \midrule
    \textbf{512}         & 0.018        & 0.032        & 0.013        & 0.051          & 0.033      \\
    \textbf{1024}        & 0.028        & 0.059        & 0.018        & 0.075          & 0.052      \\
    \textbf{2048}        & 0.054        & 0.132        & 0.053        & 0.141          & 0.112      \\ \bottomrule
    \end{tabular}}
    \caption{\label{tab:inference_time} Inference time (in seconds)}
\end{table}
\begin{table}[t]
    \setlength{\abovecaptionskip}{0.cm}
    \setlength{\belowcaptionskip}{-5pt}
    \centering
    \setlength{\tabcolsep}{3.2mm}{
    \begin{tabular}{@{}lllllll@{}}
    \toprule
    \textbf{Model}      & \textbf{MSE(R)} & \textbf{RMSE(R)} & \textbf{MAE(R)} & \textbf{MSE(t)} & \textbf{RMSE(t)} & \textbf{MAE(t)} \\ \midrule
    \textbf{DCP(SVD)}        & 550.782         & 23.468           & 17.827          & 0.0161          & 0.126            & 0.0981          \\
    \textbf{DCP(softmax)}    & 22.923          & 4.787            & 2.811           & 0.00291         & 0.0539           & 0.0345         \\ \midrule
    \textbf{OT}         & \textbf{2.526}           & \textbf{1.589}            & \textbf{0.602}           & \textbf{0.000158}        & \textbf{0.00336}          & \textbf{0.00550}          \\ \bottomrule
    \end{tabular}}
    \caption{\label{tab:Ablation 2} {Analysis of the effectiveness of our OT layer. DCP(SVD) indicates that the rigid transformation is computed within the network, as in DCP. By contrast, DCP(softmax) uses the NLL loss directly on the score map, but without relying on optimal transport to compute the correspondences. Our OT layer yields a significant accuracy boost in all the metrics.}}
\end{table}

{\bf Inference time.}
    In Table~\ref{tab:inference_time}, we compare the inference time of our method to that of the baselines for different point set sizes. For this comparison to be fair, we ran all methods on a desktop computer with an Intel(R) Core(TM) i7-7700K CPU @ 4.20GHz, an Nvidia GTX 2080 Ti GPU, and 32GB memory. For the non learning-based methods (ICP and FGR), we used their Open3D~\cite{Zhou18} implementation. For the learning-based ones (DCP and PRNet), we used Pytorch implementations. Our method yields comparable runtimes to FGR, while being slightly faster than PRNet.

{\bf Effectiveness of the OT layer.}
To evaluate the benefits of our OT layer, we compare its results with two alternatives. The first one consists of relying on the SVD layer used in DCP. Specifically, this layer takes the DGCNN features to build a score map, feeds the score map to a softmax, and then solves the Procrustes problem using SVD to obtain the rigid transformation, on which the loss function is computed. The second alternative consists of directly imposing supervision on the output of the above-mentioned softmax, i.e., avoiding the SVD computation but not relying on optimal transport to obtain the correspondences. For these experiments, we used the same setup as for partial-to-partial registration with unseen categories, except that we removed the Transformers from feature embedding network of all the methods. The reason for this modification is that, with such Transformers, training DCP(SVD) requires at least one 32GB GPU, which we don't have access to.
As shown in Table~\ref{tab:Ablation 2}, while direct supervision of the score maps (softmax) yields much more robust results than the use of an SVD layer, our OT layer yields a further significant boost in all metrics.

\section{Conclusion}

We have introduced the first one-shot learning-based partial-to-partial registration method. At the heart of our approach lies an optimal transport layer enabling our network to handle partial correspondences between the two input point sets. Our method does not require a complicated training strategy, yet yields more accurate results than other approaches that do. Furthermore, it has a reasonably low memory footprint and generalizes well to unseen data, making it broadly applicable. In the future, we will aim to deploy it for LiDAR-based registration, with a particular focus on the task of removing debris in low Earth orbit.

\section{Broader Impact}
Our work tackles the general problem of partial-to-partial 3D registration. As such, and because our approach yields high accuracy and generalizes to unseen data, many applications can potentially benefit from it. In particular, partial-to-partial registration is commonly used to reconstruct indoor scenes~\cite{Zeng17b}, with potentially high impact in the gaming and mixed reality industry. Furthermore, it can be used for self-localization in outdoor scenes, thus greatly facilitating autonomous navigation, as illustrated by the KITTI registration dataset~\cite{Geiger12}. This, we believe, can have a high societal impact, to which we will aim to actively contribute by deploying our approach in the context of space debris removal. From an ethical perspective, we do not foresee major risks, because point clouds carry very little personal information, such as gender or ethnicity. While the security risks might be higher, e.g., in the presence of adversarial attacks, we expect adversarial training to significantly mitigate them.



\section{Registration as a Procrustes Problem}
Given $n$ correspondences $(\bx_i,\by_i)$ between the source point set $\bX$ and the target point set $\bY$, our goal is to compute a rigid transformation, encoded via a rotation matrix $\bR \in SO(3)$ and a translation vector $\bt \in \mathbb{R}^3$, that aligns each $\bx_i$ to its corresponding $\by_i$. This is achieved by minimizing the mean-squared error
\begin{equation}
    \underset{\bR\in SO(3), \bt}{\mathrm{argmin}}\frac{1}{n}\sum_{i=1}^{n}\lVert\bR \bx_i + \bt - \by_i\rVert^2\;.
\end{equation}
To do so, we first discard the relative translation of the point sets by computing their center of mass
\begin{equation}
    \bbx = \frac{1}{n}\sum\limits^{n}_{i}\bx_i\;, \bby = \frac{1}{n}\sum\limits^{n}_{i=1}\by_i\;,
\end{equation}
and removing them from each point as, e.g., $\bx_i - \bbx$. Then, the rotation matrix can be obtained by minimizing the mean-squared error
\begin{align}
    \underset{\bR}{\mathrm{argmin}}&\frac{1}{n}\sum_{i=1}^{n}\lVert\bR (\bx_i - \bbx)  - (\by_i - \bby)\rVert^2 \\
    & s.t. \;\; \bR^{\top}\bR = \bI
\end{align}
which is a Procrustes problem. To solve it, we first compute the cross-covariance matrix
\begin{equation}
    \bH = \sum^{n}_{i=1}(\bx_i - \bbx)(\by_i - \bby)^{\top}\;.
\end{equation}
Then, we use singular value decomposition (SVD) to obtain $\bH = \bU\mathbf{S}\bV^{\top}$. The rotation matrix and translation vector can then be computed in closed-form as
\begin{equation}
    \bR = \bV\Sigma\bU^{\top},\;\; t = - \bR\bbx + \bby\;,
\end{equation}
where $\Sigma$ is a diagonal matrix,
where the last value 
is defined as $sign(\det(\bU\bV^{\top}))$ and the 
others are set to 1. This guarantees that the determinant of $\bR$ is positive, thus making it a valid rotation matrix. For further details please check~\cite{Schonemann66,Eggert97}.

{\small
\bibliographystyle{unsrt}
\bibliography{short,vision}
}
\end{document}